\documentclass[10pt,twocolumn,letterpaper]{article}

\usepackage{cvpr}              % To produce the CAMERA-READY version
\usepackage[dvipsnames]{xcolor}

\usepackage{graphicx}
\usepackage{booktabs}
\usepackage{nicefrac}       % compact symbols for 1/2, etc.
\usepackage{microtype}      % microtypography
\usepackage{xcolor}         % colors
\usepackage{colortbl}
\usepackage{amsmath}
\usepackage{amssymb}
\usepackage{mathtools}
\usepackage{amsthm}
\usepackage{amsmath}
\usepackage{caption}
\usepackage{subcaption}
\usepackage{enumitem}
\usepackage{makecell}
\usepackage{float}
\usepackage{placeins}
\usepackage{mathtools}
\usepackage{amsthm}
\usepackage{lineno}
\usepackage{epsfig}
\usepackage{comment}
\usepackage{multirow}
\usepackage{adjustbox}
\usepackage{textcomp}
\usepackage{xspace}
\usepackage{wrapfig}
\usepackage{setspace}
\usepackage{kantlipsum, wrapfig,graphicx}
\usepackage[rightcaption]{sidecap}
\usepackage{caption}
\usepackage{soul}
\usepackage[ruled,linesnumbered,noend,vlined]{algorithm2e}
\definecolor{cvprblue}{rgb}{0.21,0.49,0.74}
\usepackage[pagebackref,breaklinks,colorlinks,citecolor=cvprblue]{hyperref}

\def\paperID{38} % *** Enter the Paper ID here
\def\confName{CVPR}
\def\confYear{2024}

\newcommand{\divid}[0]{DIVID\xspace}

\title{Turns Out I'm Not Real: Towards Robust Detection of AI-Generated Videos}

\author{Qingyuan Liu,~~Pengyuan Shi,~~Yun-Yun Tsai,~~Chengzhi Mao,~~Junfeng Yang\\
Columbia University\\
{\tt\small \{ql2505,ps3391\}@columbia.edu~~\{yunyuntsai,mcz,junfeng\}@cs.columbia.edu} \\}

\begin{document}
\maketitle
\begin{abstract}
The impressive achievements of generative models in creating high-quality videos have raised concerns about digital integrity and privacy vulnerabilities. Recent works to combat Deepfakes videos have developed detectors that are highly accurate at identifying GAN-generated samples. However, the robustness of these detectors on diffusion-generated videos generated from video creation tools (e.g., SORA by OpenAI, Runway Gen-2, and Pika, etc.) is still unexplored. In this paper, we propose a novel framework for detecting videos synthesized from multiple state-of-the-art (SOTA) generative models, such as Stable Video Diffusion. 
We find that the SOTA methods for detecting diffusion-generated images lack robustness in identifying diffusion-generated videos. Our analysis reveals that the effectiveness of these detectors diminishes when applied to out-of-domain videos, primarily because they struggle to track the temporal features and dynamic variations between frames. To address the above-mentioned challenge, we collect a new benchmark video dataset for diffusion-generated videos using SOTA video creation tools. We extract representation within explicit knowledge from the diffusion model for video frames and train our detector with a CNN + LSTM architecture. The evaluation shows that our framework can well capture the temporal features between frames, achieves 93.7\% detection accuracy for in-domain videos, and improves the accuracy of out-domain videos by up to 16 points.

% We extract representations that imply explicit knowledge from the diffusion model for every video frame  train a CNN + LSTM detector by using the representation.
\end{abstract}    
\vspace{-3mm}
\section{Introduction}
\label{sec:intro}

% The opensource video generation tools such as Video Stable Diffusion~\cite{ho2022video}, SORA from OpenAI~\cite{videoworldsimulators2024}, Midjourney~\cite{midjourney}, Runway~\cite{runway}, and Show-1~\cite{zhang2023show} become prevalent and revolutionize industries from design, marketing, to entertainment. The high-quality video generation 

The realm of video creation is undergoing a significant transformation with the advent of video generation tools, such as Stable Video Diffusion~\cite{blattmann2023stable}, SORA by OpenAI~\cite{videoworldsimulators2024}, RunwayML~\cite{runway}, Pika~\cite{pika}, and Show-1~\cite{zhang2023show}. These cutting-edge tools are revolutionizing industries from design, marketing, and entertainment to education by creating high-quality video content. The pivotal shift is opening up a myriad of possibilities for creators everywhere, yet poses societal dangers, notably in their widespread use of spreading disinformation, propaganda, scams, and phishing -- evidenced by cases like the Taylor Swift deepfakes~\cite{taylorswift}. The potential threats underscore the importance of detecting video generated by these generative models.

\begin{figure}[t]
    \centering
    \includegraphics[width=0.85\linewidth]{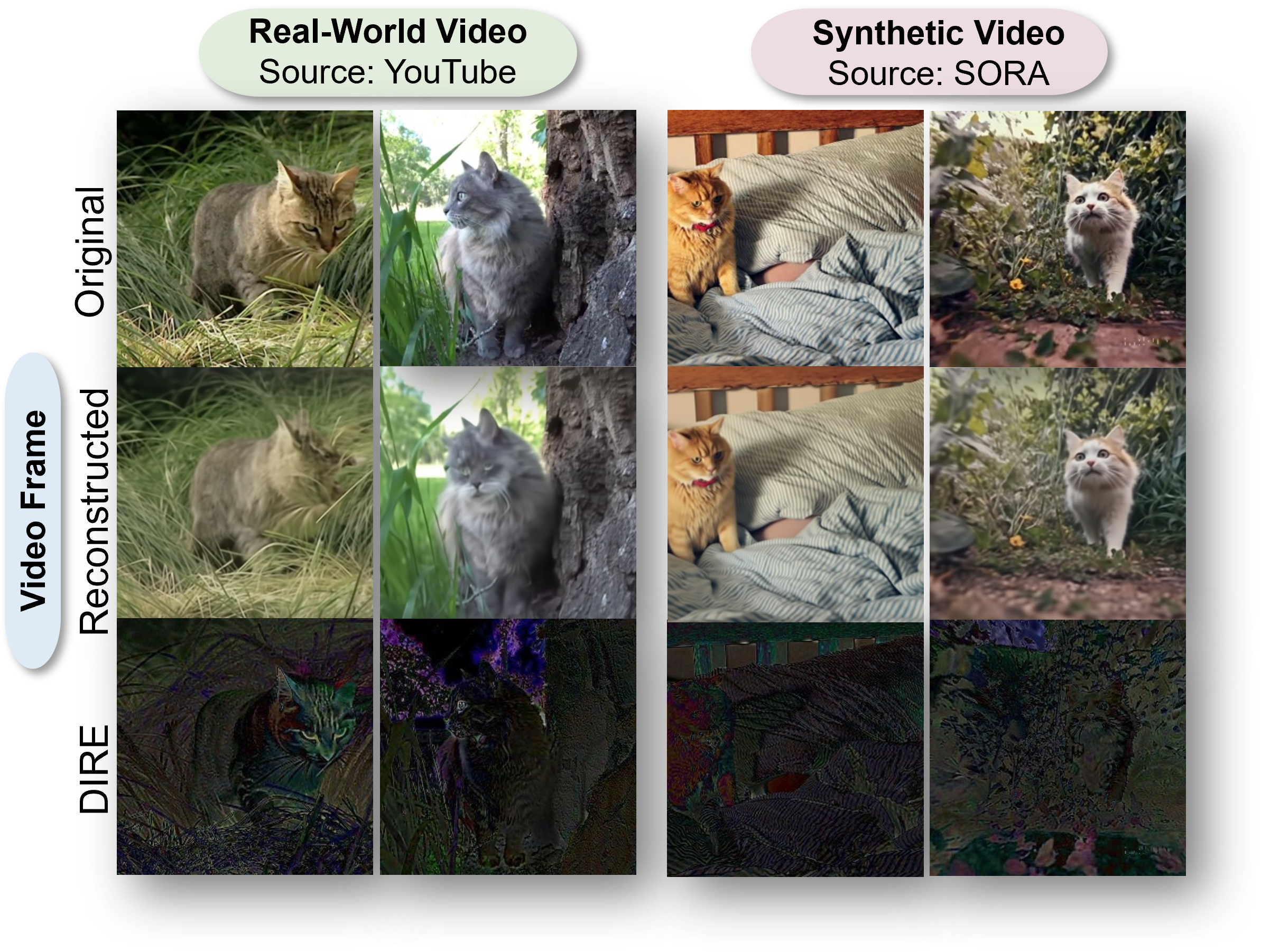}
        \vspace{-2mm}
       \caption{We show the real video frames from YouTube, and fake video from SORA by OpenAI. The explicit knowledge, DIRE, is calculated by the difference between the input original frame and the reconstructed frame from the diffusion model. 
        The reconstructed frame for the SORA-generated video will be visually close to the input original frame, yet the real video from YouTube can not (e.g., the cat face distorted after reconstruction), which inspired us to leverage the DIRE information for training.}
         \label{fig:divid_teaser}
    \vspace{-3mm}
\end{figure}
 \vspace{-1mm}
Recent advancements have developed detectors that achieve remarkable accuracy in identifying images generated by diffusion-based models~\cite{wang2023dire, ma2023exposing}. The core method first calculates the Diffusion Reconstruction Error (DIRE) by measuring the difference between an input image and its corresponding reconstructed version from the diffusion model; then, a classifier is trained on the DIRE values to distinguish the AI-generated images from human-created images. Prior work~\cite{wang2023dire} shows that these State-of-the-art detectors have great generalizability, where the detector trained with samples generated from one source of the generative model can also detect samples from other sources.
However, we found that these state-of-the-art detectors are not robust enough to detect diffusion-generated video under the same setting due to the following reasons, 
% (1) The DIRE values of multiple video frames extracted from the diffusion models can be noisy and lead to the collapse training of detectors. 
(1) The detector can easily overfit on the in-domain set when training with DIRE values only, which can not generalize to the out-domain set. (2) The SOTA diffusion-generated image detectors can not capture the temporal information in multiple video frames. (3) Existing video detection methods for deepfake are not designed for high-quality diffusion-generated video with more variance, which requires more explicit knowledge to detect.

\begin{figure}[t]
    \centering
    \includegraphics[width=0.85\linewidth]{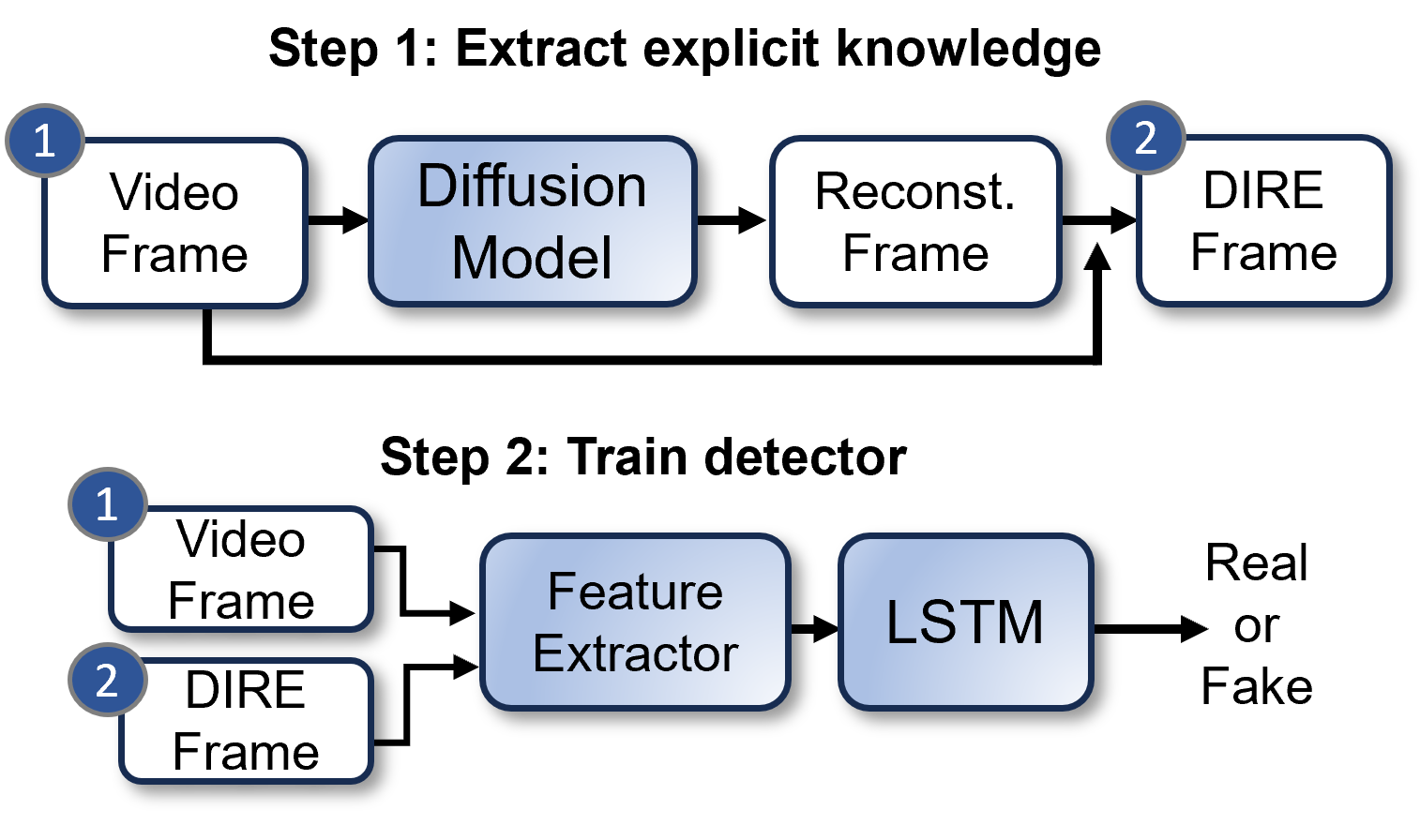}
    \label{fig:divid_flow}
        \vspace{-3mm}
       \caption{The flow of \divid. In step 1, given a sequence of video frames, we first generate the reconstructed version of every frame by using the diffusion model. Then, we calculate the DIRE values using the reconstructed frame and their corresponding input frame. In step 2, the CNN+LSTM detector is trained based on sequences of DIRE values and the original RGB frames.}
       \label{fig:divid_flow}
\vspace{-4mm}

\end{figure}

% \vspace{-1mm}
To tackle the challenges, we propose a novel approach for \textit{\textbf{DI}ffusion-generated \textbf{VI}deo \textbf{D}etection}, called \textit{\textbf{\divid}}. 
% We collect a new benchmark on diffusion-generated video using various opensource video generation tools, including Stable Video Diffusion, Pika, Runway, SORA, etc. The enchmark contains more than 1k videos. \
Our method, \textbf{\divid}, carefully investigates the sampling timestep of the diffusion process to generate DIRE values upon multiple frames for real and fake video. 
% Ideally, the diffusion-generated video can be closely reconstructed by the diffusion model, which induces small DIRE values, and the real video can not. However, 
We found the tradeoff between the quality of video reconstruction and the proper DIRE values for training the detector.
% Minimizing the sampling timestep increases noise and DIRE values for both while enlarging it will improves video's quality but lessens the DIRE gap between real and fake videos.
While the sampling step is small, the reconstruction of video frames from the diffusion model can be noisy, and the DIRE values can be large for both real and fake videos. Despite increasing the sampling timestep to a large number, which can improve the video quality after reconstruction from diffusion, it reduces the gap of DIRE values between real and fake. Therefore, a proper sampling timestep is critical for video detection.

We then propose using CNN + LSTM architectures to capture different levels of abstraction features and temporal dependencies in both original RGB frames and DIRE values, as shown in Fig~\ref{fig:divid_flow}. Prior work~\cite{wang2023dire} uses a pre-trained CNN architecture to build the detector, yet it can learn only the features of DIRE values on every single image. To incorporate the temporal features into the model, we change the model architectures and do the two-phase training. We first fine-tune the CNN on the original RGB frame and DIRE values of video, then train the LSTM network based on the feature extractor in the CNN.

Our main contributions are as follows.
\begin{itemize}
  \item We propose DIffusion-generated VIdeo Detector (\divid), a new video detection method that can detect video generated from multiple sources of video generation tools.
  \item We have collected a video dataset, including in-domain trainset / testset, and out-domain testsets. The in-domain train/test sets are from Stable Video Diffusion (SVD) model, and the out-domain is collected from Pika, Runway Gen-2, and SORA. We will release our benchmark shortly.
  \item We propose a general training framework by incorporating the temporal information meaningful explicit knowledge in video clips. We observe a simple CNN + LSTM trained with RGB frame + DIRE values can improve the generalizability of detectors on out-domain testsets.
  % \item Experimental results show that \divid achieves SOTA detection accuracy on both in-domain and out-domain video detection, compared with the baselines. 
\end{itemize}
\vspace{-1mm}

\vspace{-1mm}
\section{Related Works}
\label{sec:related_works}
\vspace{-1mm}
$\bullet$ \textbf{Diffusion-generated Video}
Diffusion-based video generation represents a leap forward from static image generation, addressing the complexities of temporal coherence, motion dynamics, and environment consistency. Tools like SORA by OpenAI~\cite{videoworldsimulators2024}, Stable Video Diffusion~\cite{blattmann2023stable}, MidJourney~\cite{midjourney}, RunwayML~\cite{runway}, Show-1~\cite{zhang2023show}, Pika~\cite{pika} and Deep Dream Generator~\cite{deepdream} enable users to generate videos with impressive visual and narrative quality. These tools illustrate the breadth of development in AI-driven video content creation, showcasing a range of capabilities from enhancing video quality to generating entirely new content. Each offers unique features tailored to specific creative or technical needs, reflecting the rapid evolution and growing accessibility of video generation technology.

$\bullet$ \textbf{Generative Video Detection}
The success in generating high-quality images has heightened concerns about security, personal privacy, and digital integrity, emphasizing the need for a robust detector to discern whether samples are from generative models.
% A large body of works has studied the detection methods for GAN-based generated images by using traditional deep neural networks. 
Recently, Deepfake video, generated by GAN-based models, can perform face manipulation with high realism~\cite{deepfake_pavel}. Agarwal et al.~\cite{deepfake_pavel} point out the challenges of detecting deepfake video, where the traditional DNN networks or audio-visual approach based on lipsync inconsistency detection are not robust enough to detect Deepfake.
Rössler et al.~\cite{rössler2019faceforensics} introduced a face forgery detection technique that begins by tracking and extracting facial information from the sample, followed by training a classifier to detect forgeries.
Marra et al.~\cite{Marra2018} developed multiple CNN-based models for detecting fake images. David et al.,~\cite{8639163_deepfake} proposed to use CNN + LSTM to do the deepfake video detection. However, their approach did not account for cross-model transferability and was found ineffective for generalizing to diffusion-based images~\cite{Corvi_2023_ICASSP, ricker2024detection}.
Lorenz et al.~\cite{Lorenz_2023_ICCV} modified the widely-used Local Intrinsic Dimensionality (LID) to detect diffusion-based images. Based on the observation that generated images vary less than real images after reconstruction. DIRE~\cite{wang2023dire} proposed to utilize reconstructed error to detect diffusion-generated images and was proved better than traditional RGB-based detectors. SeDID~\cite{ma2023exposing} extended it by ensembling step-wise noise error of corresponding inverse and reverse stages. To facilitate diffusion image detection, benchmarks at image level~\cite{wang2023dire} and fine-grained region level~\cite{wang2023deter} were established for fair comparison. Despite the prior works, the robustness of detectors on diffusion-generated video remains unexplored.

\vspace{-3mm}

\section{Method}
\label{sec:method}
\vspace{-1mm}
\subsection{Preliminaries}

$\bullet$ \textbf{Denoising Diffusion Probabilistic Models (DDPM)}
The diffusion model is a novel approach for image generation that was first proposed in~\cite{song2020denoising} inspired by nonequilibrium thermodynamics, strengthening the image quality during the generation process. 
DDPM operates by gradually transforming random noise into structured images over a series of steps, simulating a reverse diffusion process. 

Given an image $x_0$ sampled from $\mathcal{X}_S$, DDPM relies on learning the reverse of a diffusion process, where a data point $x_0$ is progressively noised until it becomes a Gaussian noise through a fixed Markov Chain for $T$ steps, and then learning to denoise it back to the original image.
Specifically, for every timestep $t \in [1, ... , T]$, they sample data sequence $[x_1, ... ,x_T]$ by adding Gaussian noise with variance $\beta_t \in (0, 1)$ during the forward process, which is defined as:
% \begin{equation}
%    q(x_{1:\mathcal{T}}|x_{0})= \Pi_{t=1}^{\mathcal{T}}(x_t | x_{t-1})`
% \end{equation}
\begin{equation}
   q(x_{t}|x_{0})= \sqrt{\bar{\alpha}_{t}} x_0 +  \sqrt{1-\bar{\alpha}_{t}} \epsilon \ ,
\label{eqn:forward}
\end{equation}
where $\epsilon\sim\mathcal{N}(0, 1)$ is the noise sampeled from Gaussian distribution, $\alpha_t = 1 - \beta_t$, and $\bar{\alpha}_{t} = \Pi_{s=1}^{t}\alpha_{s}$. 
The reverse process then generates a sequence of denoised image $[x^g_t, x^g_
{t-1} ... ,x^g_0]$ from timestep $t \in [T, ... 1]$. For timestep $t$ in the reverse process, the denoised image can be defined as:
% \begin{equation}
%    p_{\theta}(x_{0:\mathcal{T}})= p_{\theta}(x_\mathcal{T}) \Pi_{t=1}^{\mathcal{T}} p_{\theta}(x_{t-1} | x_t)
% \end{equation}
% \vspace{-2mm}
% \begin{equation}
%     p_{\theta}(x_{t-1} | x_t) = \mathcal{N}(x_{t-1}; \mu_{\theta}(x_t, t), \tilde{\beta}_t\mathbf{I}) \ ,
% \end{equation}
% where the variance $\tilde{\beta}_t = \frac{1- \bar{\alpha}_{t-1}}{1- \bar{\alpha}_t} \cdot \beta_t$ is a constant, and the mean $\mu_{\theta}(x_t, t) = \frac{1}{\sqrt{\alpha_t}} \left( x_t - \frac{\beta_t}{\sqrt{1-\bar{\alpha}_t}} \epsilon_\theta(x_t, t) \right)$ is computed by approximating the $\epsilon$ from the noise predictor $\epsilon_\theta(x_t, t)$. 
% To be simplified, the reverse process of DDPM to generate a denoised image from $x_t$ is defined as 
\begin{equation}
    x^g_{t-1} = \frac{1}{\sqrt{\alpha_t}}\left(x^g_t - \frac{1-\alpha_t}{\sqrt{1-\bar{\alpha}_t}}\epsilon_\theta(x^g_t, t)\right)+ \sigma_t \epsilon \ ,
\label{eqn:reverse}
\end{equation}
where $\epsilon_\theta$ is a trainable noise predictor that generates a prediction for the noise at the current timestep and removes the noise. $\sigma_t$ is the variance of noise. Ideally, the generated sample $x^g_0$ should be moved forward to the distribution of the source domain trained for the diffusion model.

% \paragraph{Structural Guidance in Diffusion Reverse Process}
% The trade-off between preserving content while translating domains or style has been studied by DDA~\cite{gao2022back, yang2023zeroshot}. When the noise variance $\sigma$ is more extensive, it is challenging to preserve the content information. Therefore, the structural guidance allows the diffusion model to generate samples conditioned on the predefined objectives. In particular, the structural guidance iteratively refines the latent for the input images during the reverse process so that the content information in the sample can be preserved while translating the style or shifting the domain.

% To be simplified, the reverse sampling process to generate a denoised image of $x_0$, known as $\hat{x}_{0,t}$, conditioned on the sample $x_t$ at time step $t$ is then given by:

% \begin{equation}
%     \hat{x}_{0,t} = \sqrt{\frac{1}{\bar{\alpha}_t}}x_t - \sqrt{\frac{1-\bar{\alpha}_t}{\bar{\alpha}_t}}\epsilon_\theta(x_t, t)
% \label{eqn:reverse}
% \end{equation}

% \begin{equation}
%     x_{t-1} = \sqrt{\bar{\alpha}_{t-1}}\left(x_t - \hat{x}_{0, t}\right) &+ \sqrt{1- \bar{\alpha}_{t-1}-\sigma^2}\epsilon_{\theta}(xt, t) + \sigma \epsilon
% \label{eqn:reverse}
% \end{equation}

% To adapt the model using diffusion for OOD data, we aim to generate samples by diffusion and make them close to the source domain. 

$\bullet$ \textbf{Denoising Diffusion Implicit Models (DDIM)}
% Due to the sampling process of DDPM being a Markov chain, it requires all past denoising steps to obtain the next denoised image. 
In DDPM, the long stochastic operations can lead to gradient vanishing during the training and huge distortion of the content information. Thus, to train the diffusion more efficiently without content distortion, DDIM~\cite{song2020denoising} modifies the diffusion process by introducing a non-Markovian implicit trajectory.
The reverse process in DDIM can be defined as:
\begin{align}
     x^g_{t-1} &= \sqrt{\bar{\alpha}_{t-1}}\left(x^g_t - x^g_{0, t}\right)   \\ \nonumber
     &+ \sqrt{1- \bar{\alpha}_{t-1}-\sigma_t^2}\epsilon_{\theta}(x^g_t, t) + \sigma_t \epsilon \ ,
\label{eqn:reverse_redefine}
\end{align}
where $x^g_{0,t}$ is the predicted denoised image for $x_0$ conditioned on $x^g_t$ at the time step $t$ and is defined as:
\begin{equation}   
     x^g_{0,t} = \frac{x^g_t - \sqrt{1- \bar{\alpha}_t} \epsilon_{\theta} (x^g_t, t)}{\sqrt{\bar{\alpha}_t}}, \
\label{eqn:reverse_denoise}
\vspace{-1mm}
\end{equation}
While $\sigma_t=0$, the noise term $\epsilon_\theta$ is ignored, and the sampling process becomes deterministic. 
When $\sigma_t=\sqrt{1-\Bar{\alpha}_{t-1}/ (1-\Bar{\alpha}_t)}\sqrt{1-\Bar{\alpha}_t/\Bar{\alpha}_{t-1}}$, the generative process becomes DDPM.

\subsection{Diffusion-generated Video Detector 
 (\divid)}

$\bullet$ \textbf{Diffusion Reconstruction Error (DIRE)} Previous studies indicate that image detectors designed for GAN or VAE models experience performance drops when identifying images generated by diffusion processes. Moreover, training a binary classifier solely on real and diffusion-generated images does not ensure it will effectively recognize new, unseen diffusion-generated images.
In Wang et al.,~\cite{wang2023dire}, they proposed the DIRE, which can well capture the signal of the diffusion-generated image. 
Their hypothesis posits that images from diffusion models are sampled from the diffusion process distribution, suggesting that reconstructed diffusion-generated images should closely resemble each other. For any image $x_0$ from any set $\mathcal{X}$ and a pre-trained diffusion model, the DIRE is calculated by 
\begin{equation}   
     DIRE(x_0)=\vert x_0 - \textbf{R}(\textbf{I}(x_0)) \vert \
\label{eqn:dire}
\end{equation}
where $\text{\textbf{I}}$ represents the inversion process and $\text{\textbf{R}}$ represents the reconstruction process. The output of DIRE is calculated by the absolute difference between $x_0$ and its corresponding reconstructed version. \divid leverages the DIRE to detect the diffusion-generated videos.

$\bullet$ \textbf{Detector for Diffusion-generated Videos}
To build a robust detector for detecting AI-generated videos, instead of training a binary CNN classifier with video frames, we propose a CNN+LSTM trained with RGB frames + DIRE values. The DIRE values are extracted from an unconditional diffusion model. To well learn the temporal features of RGB frames+DIRE values on every time step, we first train a CNN detector for extracting the frame features on single frames. Then, the RGB frames and DIRE values will be leveraged to train CNN + LSTM model jointly.

Assume we already extract the DIRE of the video frame set from a video sample $Q$, denoted as $\{q_1, q_2, ..., q_T\}$ and $T$ is the total time step of frame. The $\phi$ is the feature extractor of CNN model. The feature of DIRE set from $\phi$ is $\{\phi(q_1), \phi(q_2), ..., \phi(q_T)\}$. 
For a feature $\phi(q_t)$ of DIRE frame $q_t$ on time step $t \in [1,2,..., T]$, the working of LSTM in \divid can be described as follows. The LSTM generated hidden representation $a^t$ at every time step in the LSTM cells for output prediction (Eq.~\ref{eqn:lstm_eqn3}), which inquires the input feature frame $\phi(q_t)$ and the inherent information of hidden representation $a^{t-1}$ from the previous step. The cell state $c^{t}$ in Eq.~\ref{eqn:lstm_eqn2} memorizes the past cell state and helps retain the information from the past for $a^{t}$. $f^t$, $u^t$, and $o^t$ are the forgetting, updating, and output gate layers. We initially generate a candidate cell state $c^{\sim t}$ using Eq.~\ref{eqn:lstm_eqn1}, where $W^c_a$ and $W^c_x$ are the weight parameters.
\begin{equation}   
     c^{\sim t}=\text{tanh}(W^c_a a^{t-1} + W^c_x \phi(q_t)) \
\label{eqn:lstm_eqn1}
\vspace{-1mm}
\end{equation}
\begin{equation}   
     c^{t}=f^t \cdot c^{t-1} + u^t \cdot c^{\sim t} \
\label{eqn:lstm_eqn2}
\vspace{-1mm}
\end{equation}
\begin{equation}   
     a^{t}=o^{t} \cdot 
     \text{tanh}(c^{t}) \
\label{eqn:lstm_eqn3}
\vspace{-1mm}
\end{equation}
By training the CNN+LSTM with DIRE+RGB frame features, \divid improves the detection accuracy of video from both in-domain and out-domain. 
\vspace{-2mm}
% \begin{table}[t]
% \centering
% \scriptsize
% \begin{tabular}{llccc}
% \hline
%                       & \textbf{\begin{tabular}[c]{@{}c@{}}Detector \\ Architecture\end{tabular}} & \multicolumn{3}{c}{\textbf{Evaluation Metrics}}                   \\ \cline{3-5} 
%                 &      & \textbf{Acc.}  & \textbf{AUC} & \textbf{AP}   \\ \hline
% \textbf{RGB}  & \textbf{CNN}                             & 90.16          & 96.78 & 97.02          \\
% \textbf{RGB}                       & \textbf{CNN+LSTM}                & 90.16          & 97.15 & 97.39          \\
% \textbf{DIRE~\cite{wang2023dire}}         & \textbf{CNN}                            & 92.74          & 97.35 & 97.46        \\  \hline
% \textbf{\divid / DIRE only} & \textbf{CNN+LSTM}             & \textbf{93.68} & 97.31 & 97.66   \\ 
% \textbf{\divid / DIRE + RGB}  & \textbf{CNN+LSTM}            & 91.33  & \textbf{ 97.95} & \textbf{98.20}  \\ \hline
% \end{tabular}
% \vspace{-2mm}
% \caption{Detection performance on the in-domain testset. \divid outperforms our baseline architectures regarding Acc., AUC and AP. RGB represents the original pixel frame values from raw video.}
% \label{tab:in_domain}
% \vspace{-3mm}
% \end{table}

\begin{table}[t]
\centering
\scriptsize
\begin{tabular}{llcc}
\hline
                      & \textbf{\begin{tabular}[c]{@{}c@{}}Detector \\ Architecture\end{tabular}} & \multicolumn{2}{c}{\textbf{Evaluation Metrics}}                   \\ \cline{3-4} 
                &      & \textbf{Acc.}  & \textbf{AP}   \\ \hline
\textbf{RGB}  & \textbf{CNN}                             & 90.16        & 97.02          \\
\textbf{RGB}                       & \textbf{CNN+LSTM}                & 90.16           & 97.39          \\
\textbf{DIRE~\cite{wang2023dire}}         & \textbf{CNN}                            & 92.74        & 97.46        \\  \hline
\textbf{\divid / DIRE only} & \textbf{CNN+LSTM}             & \textbf{93.68}  & 97.66   \\ 
\textbf{\divid / DIRE + RGB}  & \textbf{CNN+LSTM}            & 91.33  & \textbf{98.20}  \\ \hline
\end{tabular}
\vspace{-2mm}
\caption{Detection performance on the in-domain testset. \divid outperforms our baseline architectures regarding accuracy (Acc.) and average precision (AP). RGB represents the original pixel frame values from raw video.}
\label{tab:in_domain}
\vspace{-3mm}
\end{table}

\begin{table}[t]
\centering
\scriptsize
\begin{tabular}{llcccc}
\hline
\multicolumn{1}{c}{} & \textbf{Model}  & \multicolumn{3}{c}{\textbf{Out-domain}}       & \textbf{\begin{tabular}[c]{@{}c@{}}Total \\ Avg.\end{tabular}} \\ \cline{3-5}
\multicolumn{1}{c}{} & \multicolumn{1}{c}{}                             & \textbf{Gen-2} & \textbf{Pika} & \textbf{SORA} &                                                                                \\ \hline
\textbf{RGB} & \textbf{CNN}                                                                                                           & 65.42        & 78.04        & 60.05        & 67.84                                                                         \\
\textbf{RGB} & \textbf{CNN+LSTM}                                                                                                           &\textbf{67.76}        & 84.11        &  60.80        & 70.89                                                                         \\
\textbf{DIRE~\cite{wang2023dire}}        & \textbf{CNN}                                                                                                          & 50.93           & 60.75          & 54.77          & 55.48                                                                         \\ \hline
\textbf{DIVID / DIRE only}        & \textbf{CNN+LSTM}                                                                                                       & 60.75        & 80.37        & 60.8          & 67.3                                                                         \\
\textbf{DIVID / DIRE + RGB}       & \textbf{CNN+LSTM}                                                                                                        & 66.82        & \textbf{86.92}          &  \textbf{61.01}          & \textbf{71.58}                                                                \\ \hline
\end{tabular}
\vspace{-2mm}
\caption{Detection performance on out-domain testsets. We compare \divid with three baselines and show the detection accuracies on SORA~\cite{videoworldsimulators2024}, Pika~\cite{pika}, and Gen-2~\cite{runway}.}
\label{tab:sora_result}
\vspace{-3mm}
\end{table}
% \vspace{mm}

\section{Experiment}
\label{sec:experiment}

\paragraph{\textbf{Experimental Setting}}

$\bullet$ \textbf{Dataset}
We contruct a dataset for evaluating our method shown in Table~\ref{tab:dataset_detail} by using public video generation tools, including Stable Video Diffusion (SVD)~\cite{ho2022video}, Pika~\cite{pika}, Gen-2~\cite{runway}, SORA~\cite{videoworldsimulators2024}. 
Our source data are from ImageNet Video Visual Relation Detection (VidVRD)~\cite{shang2017video}, which contains 1k source videos. 
% For each video, we use (1.) SVD conditioned on all the clips to generate our in-domain dataset. (2.) Pika and Gen-2 conditioned on the test portion of the clips to generate our out-domain dataset. 
For fake video generation with SVD, the process starts from randomly cropping each source video into a 25-frame video clip. Then, the first image frame will be sent to SVD to generate a corresponding 25-frame fake video clip. 
% Finally, the first frame is dropped for both real and fake video clips, resulted in a real and fake video clip pair of 24-frame length each. 
The video resolution is 1025$\times$576.
% For Pika~\cite{pika} and Gen-2~\cite{runway} benchmarks, we use the validation set from Imagenet-VidVRD, as the conditions to generate fake videos. We collect 107 fake videos for Pika and Gen-2 separately. 
% For Pika~\cite{pika} and Gen-2~\cite{runway} benchmarks, we generate 1 fake video conditioned on each source video in the validation set of from Imagenet-VidVR (107 videos).
% The real video dataset is from ImageNet Video Visual Relation Detection (VidVRD)~\cite{shang2017video} 
% which contains 1k videos selected from ILVSRC2016-VID~\cite{ILSVRC15} dataset. 
% To ensure the generation quality, we used the same resolution as train set of SVD which is 1024x576, and we generated one second video at 25 fps. Under this configuration, we generated 2k fake videos based on the input of first and middle frame of corresponding real videos. At the same time, we capture the first and middle 25 frames to serve as real videos dataset.
% For PIKA and Gen-2, same as SVD, we provide the first and middle frame of real video as input and generated the fake video. .
 % For SORA benchmark, we select 34 videos from the opensource webpage, and collected 3 to 5 real videos of corresponding SORA video from YouTube with the same theme (e.g., A cat on the bed).
% The fake video from SVD and real video from ImageNet-VidVRD are served as the in-domain dataset for \divid to train and evaluate our detector. In addition, we evaluate \divid on three out-domain video set, including PIKA, Gen-2, and SORA. We show the results for both in-domain and out-domain in the experimental results.
$\bullet$  \textbf{Model} We use ADM~\cite{dhariwal2021diffusion}, an unconditional 256$\times$256 diffusion model trained on ImageNet-1K~\cite{ILSVRC15}, as our reconstruction model to generate the DIRE representation for every video frame. The CNN classifier is a ResNet50 model with pre-trained weight on ImageNet-1K. The LSTM follows a one-layer architecture with hidden size 2048, which can handle multiple types of representation (e.g., DIRE, RGB values of the original frame) extracted from the CNN feature extractor.
$\bullet$ \textbf{Baseline and Implementation Details} \divid is trained based on DIRE and original RGB features extracted from video frames with a CNN + LSTM model. We set the training batchsize as 128. In the training of LSTM, we use a 32-consecutive sequence of 4 frames in each batch. 
% Then we compare \divid with the detector trained with DIRE or original RGB using CNN with ResNet50.
$\bullet$ \textbf{Evaluation Metrics} We evaluate the detection performance based on the prediction of every frame and calculate accuracy and average precision.

\begin{table}[t]
\centering
\scriptsize
\begin{tabular}{ccccc}
\hline
\multicolumn{1}{l}{}                                                            & \begin{tabular}[c]{@{}c@{}}Video \\ Source\end{tabular} & \begin{tabular}[c]{@{}c@{}}Denoising \\ Condition\end{tabular} & \begin{tabular}[c]{@{}c@{}}Generated \\ Model\end{tabular} & \begin{tabular}[c]{@{}c@{}}\# of  Clips \\ (real/fake)\end{tabular} \\ \hline
\begin{tabular}[c]{@{}c@{}}In-\\ domain\end{tabular}                            & VidVRD~\cite{shang2017video}                                                           &Image2Video                                                    & SVD-XT~\cite{ho2022video}                                                     & 1k/1k                                                      \\ \hline
\multirow{3}{*}{\begin{tabular}[c]{@{}c@{}}Out-\\ domain\end{tabular}} & VidVRD (test)                                                 & Image2Video                                                             & Pika~\cite{pika}                                                                & 107/107                                                             \\
                                                                                & VidVRD (test)                                               & Image2Video                                                             & Gen-2~\cite{runway}                                                               & 107/107                                                             \\
                                                                                & YouTube/SORA~\cite{videoworldsimulators2024}                                            & x                                                                       & x                                                                   & 207/191                                                             \\ \hline
\end{tabular}
\vspace{-2mm}
\caption{Composition of our video dataset. It includes real video clips from VidVRD~\cite{shang2017video} and fake video clips generated from SVD, Pika, and Gen-2. The fake video generation is conditioned on the video frame from VidVRD. We also collected real videos from YouTube and SORA~\cite{videoworldsimulators2024} website as our 3rd out-domain test set based on the same them (e.g., A cat on the bed).}
\label{tab:dataset_detail}
\end{table}
    \vspace{-2mm}

\begin{figure}[h]
    \centering
    \includegraphics[width=0.99\linewidth]{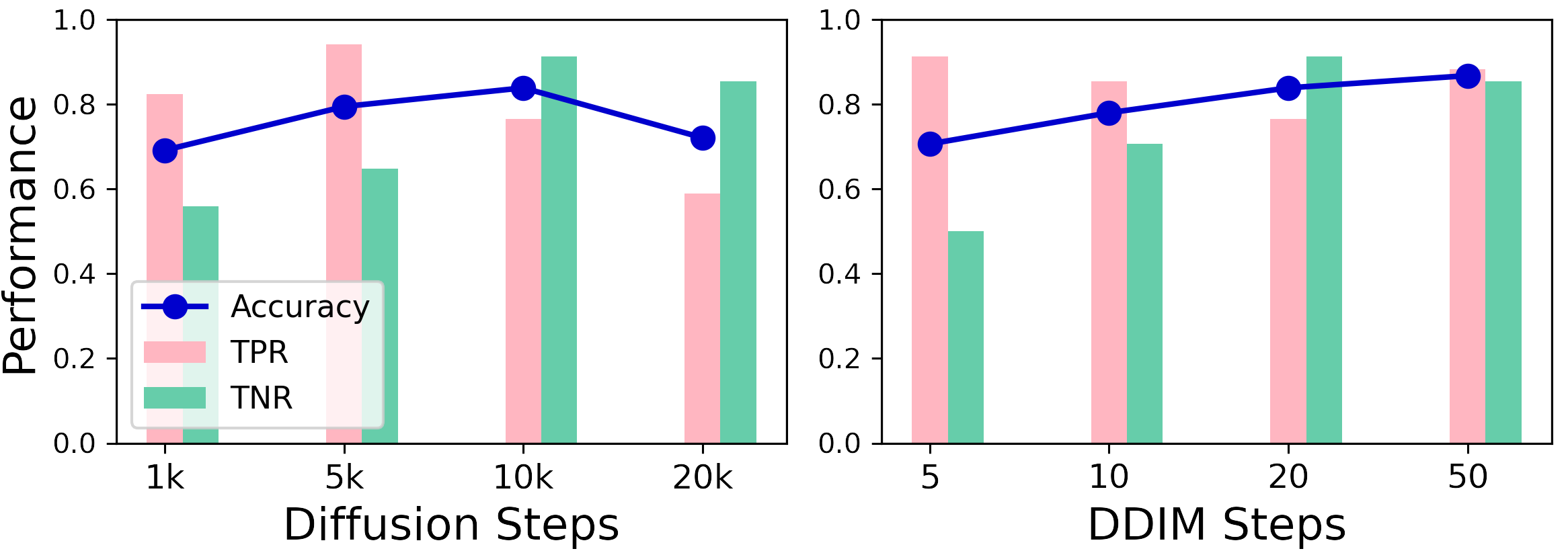}

    \vspace{-3mm}
    \caption{Analysis on diffusion steps and ddim step for \divid. The left-side subfigure shows the performance on different diffusion steps from 1k to 10k, we freeze the ddim step as 20 for all. The right-side subfigure shows the performance on different ddim steps from 5 to 50, and the diffusion step is fixed as 10k for all.}
        \label{fig:step_analysis}
 \vspace{-4mm}
\end{figure}

\vspace{-1mm}
\paragraph{\textbf{Experimental Results}}
We show the detection performance for in-domain video set in Table~\ref{tab:in_domain}. The three baselines, including original trained under CNN and CNN+LSTM, and DIRE~\cite{wang2023dire} trained with CNN. Compared with the three baselines, \divid achieves 98.20\% average precision (AP) and has better detection accuracy, and outperforms them by 0.94\% to 3.52\%. In Table~\ref{tab:sora_result}, we show the performance of \divid on three out-domain testsets, including SORA, Pika, and Gen-2. For every baseline, we show the checkpoint results that have the best out-domain detection accuracy. We observe that \divid has competitive in-domain results as baselines and also achieves better generalizability on out-domain testset. \divid improves the out-domain average accuracy by  0.69\% to 16.1\%. In Figure~\ref{fig:step_analysis}. we compare different diffusion steps and DDIM steps for generating the DIRE values in our \divid.

\vspace{-2mm}
\section{Conclusion}
\label{sec:conclusion}

We propose a general framework, \divid for diffusion-generated video detection. We have collected a new video benchmark dataset, including fake videos generated from SVD, SORA, Pika, and Gen-2. Unlike prior SOTA detectors that only use the DIRE values to train the CNN detector, \divid leverages both RGB frames and DIRE values with a simple CNN+LSTM, which can capture the temporal information and extract explicit knowledge from multiple video frames. The evaluation shows that \divid achieves better detection performance on in- and out- domain testsets by up to 3.52 and 16.1 points. Our work highlights the importance of increasing the generalizability of current SOTA detectors. 
% Future directions include studying more explicit knowledge in video frames and efficiently learning the detector.
\clearpage
{
    \small
    \bibliographystyle{ieeenat_fullname}
    \bibliography{main}
}
\end{document}